\documentclass[sigconf,nonacm]{acmart}
\usepackage{csquotes}
\usepackage{booktabs} 
\usepackage{multirow}
\usepackage{graphicx} 
\usepackage{subcaption}  
\usepackage{float}

\AtBeginDocument{%
  }

\setcopyright{acmlicensed}
\copyrightyear{2026}
\acmYear{2026}
\acmDOI{XXXXXXX.XXXXXXX}
\acmConference[Conference '26]{Make sure to enter the correct
  conference title from your rights confirmation email}{April,
  2026}{Woodstock, NY}
\acmISBN{978-1-4503-XXXX-X/2026/06}

\begin{document}

\title{Decouple and Rectify: Semantics-Preserving Structural Enhancement for Open-Vocabulary Remote Sensing Segmentation}

\author{Jie Feng}
\email{jiefeng0109@163.com}
\affiliation{%
  \institution{Xidian University}
  \country{China}
}

\author{Fengze Li}
\email{fengzeli@stu.xidian.edu.cn}
\affiliation{%
  \institution{Xidian University}
  \country{China}
}

\author{Junpeng Zhang}
\authornote{Corresponding author.}
\email{junpengzhang@xidian.du.cn}
\affiliation{%
  \institution{Xidian University}
  \country{China}
}

\author{Siyu Chen}
\email{chensy@ieee.org}
\affiliation{%
 \institution{Jimei University}
 \country{China}}

\author{Yuping Liang}
\email{yupingliang@stu.xidian.edu.cn}
\affiliation{%
  \institution{Xidian University}
  \country{China}}

\author{Junying Chen}
\email{jychense@scut.edu.cn}
\affiliation{%
  \institution{South China University of Technology}
  \country{China}}

\author{Ronghua Shang}
\email{rhshang@mail.xidian.edu.cn}
\affiliation{%
  \institution{Xidian University}
  \country{China}
}

\renewcommand{\shortauthors}{Jie Feng, et al.}

\begin{abstract}
Open-vocabulary semantic segmentation in the remote sensing (RS) field requires both language-aligned recognition and fine-grained spatial delineation. Although CLIP offers robust semantic generalization, its global-aligned visual representations inherently struggle to capture structural details. Recent methods attempt to compensate for this by introducing RS-pretrained DINO features. However, these methods treat CLIP representations as a monolithic semantic space and cannot localize where structural enhancement is required, failing to effectively delineate boundaries while risking the disruption of CLIP’s semantic integrity. To address this limitation, we propose DR-Seg, a novel decouple-and-rectify framework in this paper. Our method is motivated by the key observation that CLIP feature channels exhibit distinct functional heterogeneity rather than forming a uniform semantic space. Building on this insight, DR-Seg decouples CLIP features into semantics-dominated and structure-dominated subspaces, enabling targeted structural enhancement by DINO without distorting language-aligned semantics. Subsequently, a prior-driven graph rectification module injects high-fidelity structural priors under DINO guidance to form a refined branch, while an uncertainty-guided adaptive fusion module dynamically integrates this refined branch with the original CLIP branch for final prediction. Comprehensive experiments across eight benchmarks demonstrate that DR-Seg establishes a new state-of-the-art.
\end{abstract}


\maketitle

\section{Introduction}

\begin{figure}[t]
    \centering
    \subfloat[Existing methods]{\includegraphics[width=0.45\linewidth]{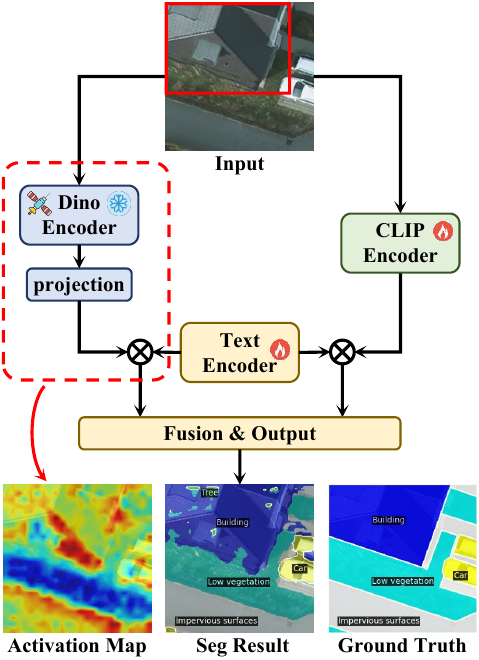}
    \label{fig:paradigm_existing_methods}}
    \hfill
    \vrule
    \hfill
    \subfloat[Our method]{\includegraphics[width=0.45\linewidth]
    {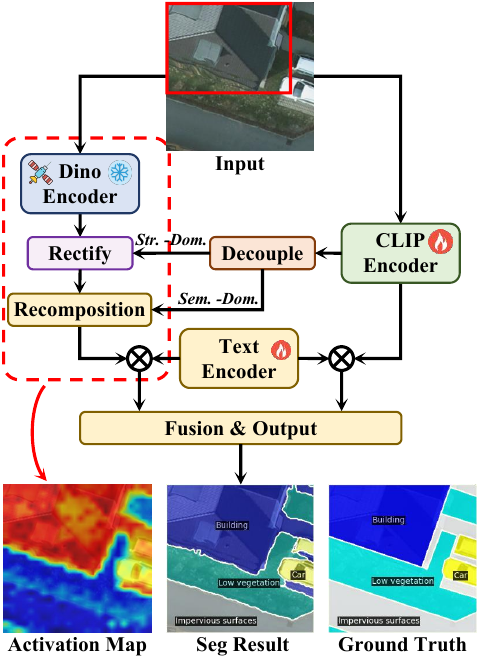}
    \label{fig:paradigm_our_methods}}
    \caption{Comparison of DINO fusion paradigms in RS open-vocabulary semantic segmentation. (a) Existing methods apply non-selective structural enhancement, resulting in diffused responses and coarse boundaries. (b) DR-Seg decouples the representation and selectively injects structural priors into the structure-dominated subspace, producing sharper boundaries while preserving semantic integrity.}
\end{figure}

Remote sensing semantic segmentation is crucial for Earth observation applications such as urban planning, land cover mapping, and disaster assessment~\cite{lu2016joint, 9877906, ru2021land, li2022dkdfn, garshasbi2025uncertainty}. However, most existing methods remain confined to a closed set of training categories, which conflicts with the open-world nature of remote sensing scenes and motivates open-vocabulary segmentation for unseen classes.

The recent success of vision-language models, particularly CLIP~\cite{clip}, has significantly propelled open-vocabulary segmentation in natural images by aligning visual and textual embeddings in a shared space, enabling flexible recognition via text prompts. Nevertheless, directly adapting CLIP-based approaches~\cite{catseg, sed} to RS imagery remains challenging. RS images are captured from an overhead perspective and typically contain densely distributed objects amid complex backgrounds, characteristics that differ markedly from natural scenes. While CLIP excels at image-level semantic alignment, its features often fail to preserve the spatial fidelity required for dense prediction, limiting its ability to model fine-grained structural details in high-resolution RS images.

To enhance structural perception, recent RS open-vocabulary segmentation methods often introduce RS-pretrained DINO as auxiliary geometric priors~\cite{GSNet, RSKT-seg}. Yet, these approaches typically exploit DINO features in a straightforward manner. As illustrated in Fig.~\ref{fig:paradigm_existing_methods}, DINO features are first projected into a CLIP-compatible space and then used to compute a correlation map with text embeddings, which is holistically fused with the corresponding features from the CLIP branch. 
However, as shown in the same figure, rather than enhancing the structural details of the activation maps, such straightforward fusion inadvertently corrupts the original semantic integrity. Fundamentally, this direct integration approach relies on an implicit assumption: it treats CLIP features as a uniform semantic space, presuming that external structural cues can be injected indiscriminately. However, our empirical observations demonstrate that this premise frequently does not hold in practice.

\begin{figure}[t]
    \centering
    \includegraphics[width=\linewidth]{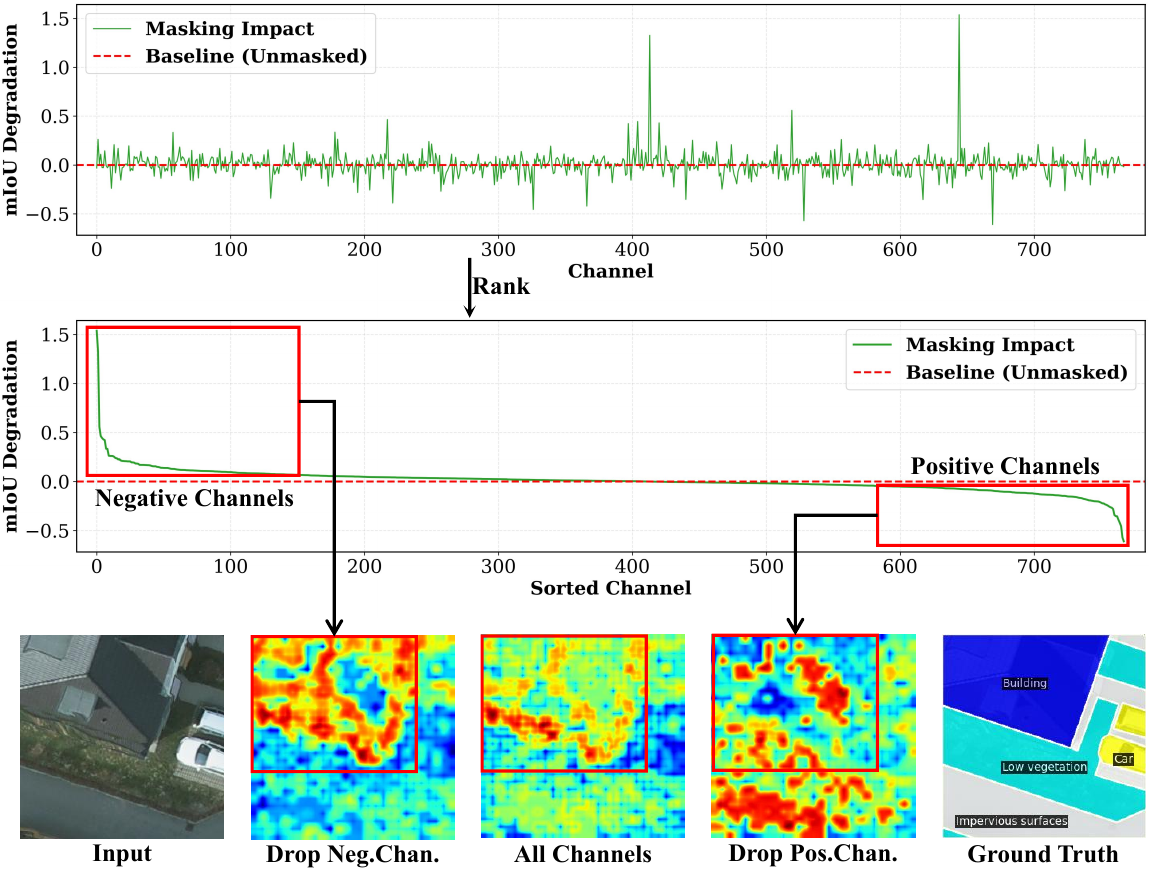}
    \caption{Channel-wise impact analysis of CLIP representations on Potsdam. \textbf{Top \& Middle:} The mIoU drop is highly uneven when individual channels are independently masked. By ranking the channels based on these drops, we categorize them into "positive" and "negative". \textbf{Bottom:} Activation maps further reveal distinct channel roles: dropping the top 20\% negative channels sharpens spatial responses, whereas dropping the top 20\% positive ones severely weakens discriminative activations.}
    \label{fig:channel_analysis}
\end{figure}

We conduct a channel-wise pilot study using 200 randomly sampled images from the Potsdam dataset. By iteratively masking each visual channel along with its corresponding text dimension, we observe highly uneven impacts on segmentation performance (Fig.~\ref{fig:channel_analysis}): some channels cause significant mIoU drops ("positive channels"), while others show minimal effect or may even improve performance ("negative channels"). Activation maps further corroborate this functional heterogeneity: dropping negative channels yields sharper and more localized responses, whereas dropping positive channels severely weakens discriminative activations. Taken together, these observations suggest that CLIP is not a homogeneous semantic space, but rather a composition of functionally heterogeneous channels. More examples are provided in the supplementary material. Existing DINO integration methods overlook this heterogeneity, thereby applying structural enhancement non-selectively and potentially disrupting the language-aligned semantics preserved by CLIP.

Motivated by the above observation, we propose \textbf{DR-Seg}, a novel \textbf{decouple-and-rectify} framework for open-vocabulary remote sensing segmentation. As shown in Fig.~\ref{fig:paradigm_our_methods}, DR-Seg retains the original CLIP branch to preserve language-aligned semantics, while constructing a refined branch for selective structural enhancement. Specifically, CLIP representations are first decoupled into a \emph{semantics-dominated subspace} and a \emph{structure-dominated subspace} based on channel-wise semantic selectivity, characterized by prototype-level cross-class activation entropy and inter-class similarity. A \textit{Prior-Driven Graph Rectification} module then injects DINO-guided structural priors into the structure-dominated subspace through graph rectification. The rectified structural representation is recomposed with the preserved semantics-dominated representation to form the refined branch, which is finally integrated with the original CLIP branch by an \textit{Uncertainty-Guided Adaptive Fusion} module according to pixel-wise prediction uncertainty. In this way, DR-Seg achieves semantics-preserving structural enhancement for open-vocabulary remote sensing segmentation.

The main contributions of this paper are summarized as follows:
\begin{itemize}
\item We identify a critical limitation of existing methods for leveraging structural priors in CLIP-based open-vocabulary segmentation, revealing clear \emph{functional heterogeneity} in CLIP representations, which provides a key motivation for selective structural enhancement.

\item We propose \textbf{DR-Seg}, a decouple-and-rectify framework for remote sensing open-vocabulary segmentation that achieves \emph{semantics-preserving structural enhancement} by decoupling CLIP features into semantics-dominated and structure-dominated subspaces, rectifying the latter with DINO guidance, and adaptively fusing the refined branch with the original CLIP branch.

\item Extensive experiments on eight remote sensing benchmarks demonstrate that DR-Seg establishes new state-of-the-art performance, with average mIoU improvements of +2.83\% on DLRSD training setting and +3.86\% on iSAID training setting.

\end{itemize}
 
\section{Related Work}
\subsection{Open-Vocabulary Semantic Segmentation}
Open-vocabulary semantic segmentation (OVS) extends semantic segmentation to an open label space specified by text prompts, largely enabled by vision-language models such as CLIP~\cite{clip}. Existing methods typically follow two main paradigms: training-free dense inference and training-based dense alignment.

Training-free methods directly reuse frozen vision-language models for pixel-level prediction by refining CLIP features at inference time. Representative works such as MaskCLIP~\cite{maskclip} and SC-CLIP~\cite{sc-clip}, together with subsequent calibration and affinity-refinement methods~\cite{clearclip, sclip, proxyclip, naclip, resclip, cass, corrclip}, improve dense prediction through token calibration, attention adjustment, or local consistency enhancement. Although efficient, these methods remain limited by CLIP's weak spatial inductive bias inherited from image-level pretraining.

Fine-tuning-based methods, in contrast, fine-tune pre-trained models via parameter updates to achieve stronger dense visual-semantic alignment. Representative approaches include mask-based methods such as ZegFormer~\cite{zegformer}, OVSeg~\cite{ovseg}, OpenSeg~\cite{openseg}, and SimBaseline~\cite{simbaseline}, as well as dense alignment methods such as LSeg~\cite{lseg}, SAN~\cite{san}, and ODISE~\cite{odise}. More recently, cost-aggregation frameworks such as CAT-Seg~\cite{catseg} and SED~\cite{sed} have become strong solutions for dense OVS, and ESC-Net~\cite{escnet} further introduces SAM\cite{sam} priors to improve boundary quality. 

\subsection{Open-Vocabulary Remote Sensing Semantic Segmentation}
While general OVS methods have achieved strong progress on natural images, extending them to remote sensing imagery is non-trivial due to overhead viewpoints, arbitrary object orientations, dense spatial layouts, and large-scale variations. Existing studies mainly improve RS open-vocabulary segmentation either by enhancing frozen foundation models at inference time or by incorporating RS-specific semantic and structural priors.

Representative of the first category, the SegEarth-OV series~\cite{segearthov, segearthov2, segearthov3} adapts frozen foundation models to RS imagery by restoring spatial details from coarse representations and denoising predictions. However, these methods are fundamentally limited by frozen features and the inherent lack of explicit structural inductive biases.

Another line introduces RS-specific priors to improve semantic discrimination or geometric robustness. SCORE~\cite{score} enhances category embeddings with scene context, while OVRS~\cite{OVSNet} addresses orientation variation through rotation-aggregative similarity computation. More closely related to our work, GSNet~\cite{GSNet} and RSKT-Seg~\cite{RSKT-seg} introduce RS-pretrained DINO guidance to strengthen CLIP-based prediction. These methods verify the value of RS structural priors, but their enhancement is still largely performed holistically in the correlation map, without explicitly identifying which specific CLIP components actually require structural enhancement. This limitation is particularly critical in remote sensing imagery, where dense layouts and fine-grained boundaries place stronger demands on structural precision. Different from these approaches, our method treats CLIP features as internally heterogeneous and performs selective structural enhancement rather than holistic refinement. This design better preserves semantic alignment while improving structural precision in open-vocabulary remote sensing segmentation.

\section{Proposed Method}

\begin{figure*}[t]
    \centering
    \includegraphics[width=\textwidth]{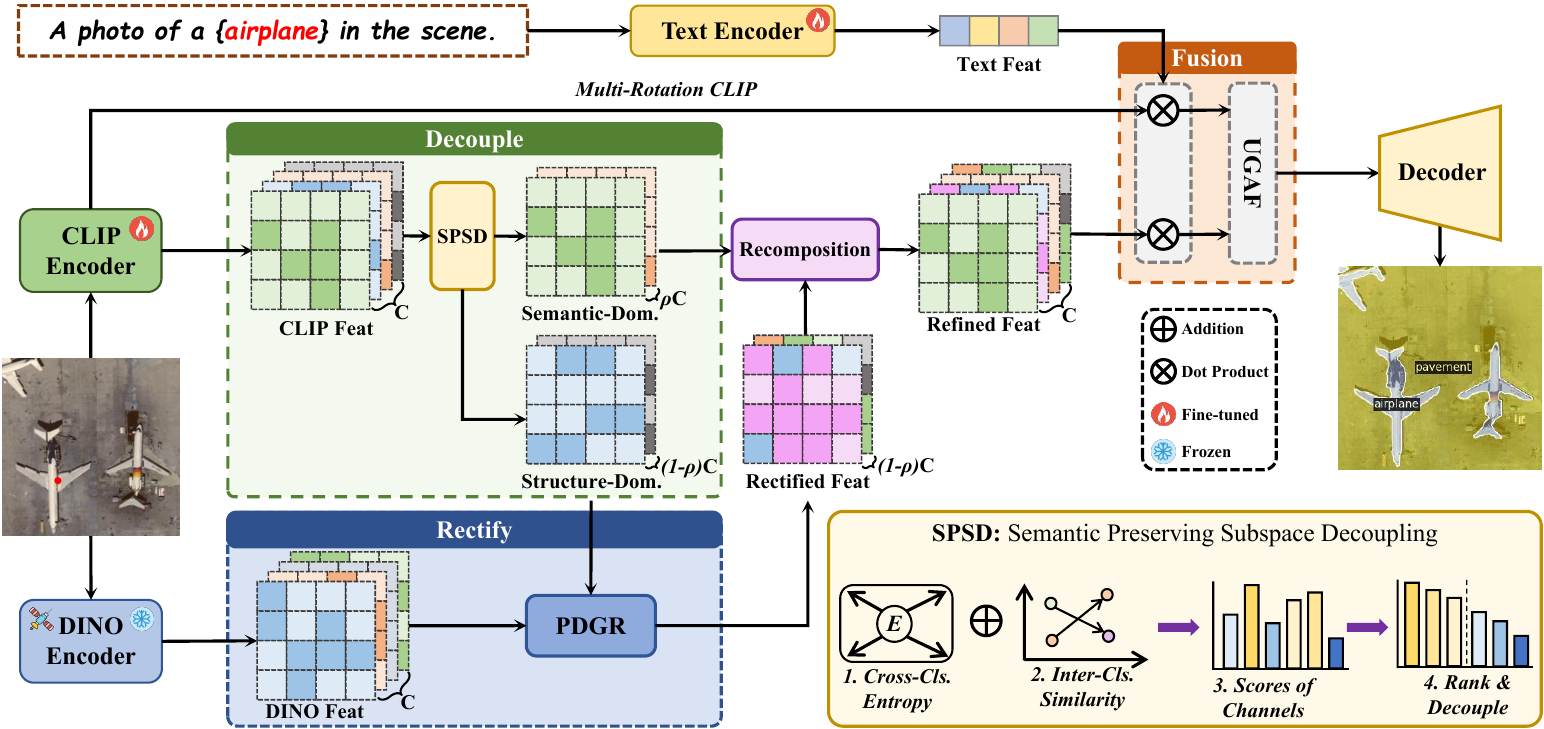}
    \caption{Overall framework of DR-Seg. The proposed pipeline consists of three stages: \textbf{Decouple}, which uses SPSD separating CLIP features into semantics-dominated and structure-dominated subspaces via SPSD; \textbf{Rectify}, which selectively rectifies the structure-dominated subspace under DINO structural guidance through PDGR; and \textbf{Fusion}, which adaptively integrates the original multi-rotation CLIP branch and the refined branch via UGAF for final segmentation prediction.}
    \label{fig:framework}
\end{figure*}

\subsection{Overall Framework}
Given a remote sensing image $I \in \mathbb{R}^{H \times W \times 3}$ and a set of category-specific textual descriptions $\mathcal{T}=\{t^{(i)}\}_{i=1}^{N_c}$, open-vocabulary remote sensing segmentation aims to assign each pixel in $I$ to the most semantically relevant category in $\mathcal{T}$, including categories unseen during training. Accordingly, DR-Seg follows the overall pipeline illustrated in Fig.~\ref{fig:framework}. Specifically, we employ a fine-tuned CLIP visual encoder and text encoder to extract visual and textual representations, while a frozen RS-pretrained DINO encoder provides structure-aware features. DR-Seg then performs semantics-preserving structural enhancement through three sequential components: \textbf{Semantic-Preserving Subspace Decoupling (SPSD)}, which partitions CLIP channels into a semantics-dominated subset and a structure-dominated subset; \textbf{Prior-Driven Graph Rectification (PDGR)}, which rectifies the latter subset under DINO guidance to produce a refined visual representation; and \textbf{Uncertainty-Guided Adaptive Fusion (UGAF)}, which adaptively fuses the original and refined predictions for final segmentation.

\subsection{Semantic-Preserving Subspace Decoupling}

While recent studies~\cite{declip, zori} attempt to mitigate semantic degradation during dense adaptation, they typically rely on self-attention modifications or simple parameter-freezing. Crucially, they lack explicit feature-level routing for selective structural enhancement. By contrast, leveraging the distinct functional heterogeneity of CLIP channels, our \textbf{Semantic-Preserving Subspace Decoupling} module offers a more direct solution. Before introducing DINO guidance, SPSD explicitly partitions features into a \emph{semantics-dominated subspace} (preserved for language recognition) and a \emph{structure-dominated subspace} (routed for structural enhancement). For stable and interpretable decomposition, this partition is computed offline on the training set(DLRSD or iSAID) and fixed throughout training and inference.

Specifically, we estimate the semantic importance of each CLIP channel using class-wise feature prototypes extracted by the frozen CLIP visual encoder. For each class $k$, we collect normalized pixel features from the corresponding ground-truth regions and sample $N_p$ prototypes to form a prototype set $\mathcal{P}_k \in \mathbb{R}^{N_p \times C}$. Using these prototypes, we evaluate the semantic importance of each channel by assessing how selectively it responds to individual classes and how effectively it distinguishes between them.

To quantify class selectivity, we assume that a semantically informative channel should respond selectively to specific categories rather than activating uniformly across all classes. Let $u_{k,c}$ denote the mean ReLU activation of channel $c$ for class $k$. We first normalize the class-wise responses into a probability distribution as:
\begin{equation}
p_{k,c} = \frac{u_{k,c}}{\sum_{j=1}^{N_c} u_{j,c} + \epsilon},
\end{equation}
and then compute the entropy of channel $c$ as:
\begin{equation}
\mathcal{H}_c = - \sum_{k=1}^{N_c} p_{k,c} \log_2 (p_{k,c} + \epsilon),
\label{eq:spsd_entropy}
\end{equation}
where a lower $\mathcal{H}_c$ indicates that the response of channel $c$ is concentrated on fewer categories, and thus reflects stronger semantic selectivity.

As a channel might still produce highly similar responses for prototypes from different classes, we further evaluate whether a channel can separate categories at the prototype level. Let $\Omega$ denote the set of prototype pairs from different classes, and let $v_{i,c}$ be the scalar response of channel $c$ for prototype $v_i$. We define the inter-class similarity of channel $c$ as:
\begin{equation}
\mathcal{S}_c = \frac{1}{|\Omega|} \sum_{(i,j)\in\Omega} (v_{i,c} \cdot v_{j,c}),
\label{eq:spsd_similarity}
\end{equation}
where a lower $\mathcal{S}_c$ indicates that channel $c$ produces less similar responses for prototypes from different classes, and therefore has better semantic discriminability.

Finally, to comprehensively determine the semantic importance of each channel, we integrate these two complementary criteria. Specifically, we rank all channels according to $\mathcal{H}_c$ and $\mathcal{S}_c$ respectively, normalize their ranks to the range $[0, 1]$, and compute the final semantic score for channel $c$ as:
\begin{equation}
\text{Score}_c = \lambda \big(1-\mathrm{Rank}(\mathcal{H}_c)\big) + (1-\lambda)\big(1-\mathrm{Rank}(\mathcal{S}_c)\big),
\label{eq:spsd_score}
\end{equation}
where $\lambda$ balances the two criteria, a larger score indicates that the channel is more important for preserving semantic alignment.

After sorting channels in descending order according to $\text{Score}_c$, we use a semantic ratio $\rho \in [0,1]$ to partition the CLIP feature channels into two index subsets:
\begin{equation}
\begin{split}
\mathcal{I}_{sem} = \{\pi_0,\dots,\pi_{\lfloor \rho C-1 \rfloor}\}, \\
\mathcal{I}_{str} = \{\pi_{\lfloor \rho C \rfloor},\dots,\pi_{C-1}\},
\end{split}
\label{eq:spsd_partition}
\end{equation}
where $\pi$ denotes the channel permutation after sorting. Given the CLIP feature $F_{\text{clip}}$, we obtain the two subspaces by channel slicing:
\begin{equation}
\begin{split}
F_{sem} &= F_{\text{clip}}[\cdot,\mathcal{I}_{sem}] \in \mathbb{R}^{h \times w \times C_{sem}}, \\
F_{str} &= F_{\text{clip}}[\cdot,\mathcal{I}_{str}] \in \mathbb{R}^{h \times w \times C_{str}},
\end{split}
\label{eq:spsd_split}
\end{equation}
where $C_{sem}=\lfloor \rho C \rfloor$ and $C_{str}=C-C_{sem}$.

The resulting decoupling explicitly preserves the semantically critical component of CLIP while exposing the remaining structure-dominated component for subsequent rectification.

\subsection{Prior-Driven Graph Rectification}

\begin{figure}[!t]
    \centering
    \includegraphics[width=\linewidth]{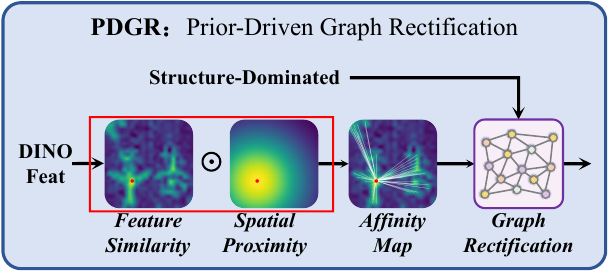}
    \caption{Detailed illustration of PDGR. It constructs a sparse affinity map using DINO-based feature similarity and spatial proximity. The resulting sparse graph, visualized by white lines (top-$k$ neighbors), dynamically rectifies the structure-dominated subspace via prior-driven spatial routing.}
    \label{fig:pdgr}
\end{figure}

Although the semantics-dominated subspace preserves language alignment, the structure-dominated subspace ($F_{str}$) requires further enhancement to capture fine-grained boundaries. Unlike recent methods~\cite {lposs} that rely on training-free, post-matching logit refinement lacking domain adaptability, our structural enhancement is intrinsically learnable and occurs before text matching. We directly rectify $F_{str}$ using a parameterized graph module. This allows the network to actively adapt DINO's structural priors to the sparse and irregular geometric layouts of remote sensing imagery, overcoming the limitations of standard convolutions or dense attention. Specifically, we construct a DINO-guided sparse affinity graph over $F_{str}$ to perform learnable rectification among structurally consistent locations, as depicted in Fig.~\ref{fig:pdgr}.

To construct the affinity graph, the affinity $A_{ij}$ between any two nodes $i$ and $j$ is jointly determined by their DINO feature similarity and spatial proximity. Formally, given $N = h \times w$ spatial locations, we denote the DINO feature vector and spatial coordinate for the $i$-th location as $f_{\text{dino}}^i \in \mathbb{R}^{D}$ and $c_i \in \mathbb{R}^{2}$, respectively. With $d_{\max}$ representing the maximum pairwise spatial distance, $A_{ij}$ is computed as:

\begin{equation}
A_{ij} = \sigma_f \cdot
\left(
\frac{f_{\text{dino}}^i \cdot f_{\text{dino}}^j}
{\|f_{\text{dino}}^i\| \, \|f_{\text{dino}}^j\|}
\right)^2
\cdot
\exp\left(
-\sigma_s \cdot \frac{\|c_i-c_j\|^2}{d_{\max}+\epsilon}
\right),
\label{eq:pdgr_affinity}
\end{equation}
where $\sigma_f$ is a scaling factor and $\sigma_s$ controls the strength of spatial decay. This affinity is used to determine graph connectivity, and we retain only the top-$k$ neighbors for each node to construct a sparse graph.

For each retained edge $(i,j)$, we further predict a learnable propagation weight from the paired DINO features:
\begin{equation}
w_{ij} =
\zeta\!\left(
\mathcal{F}_{\text{MLP}}\!\left([f_{\text{dino}}^i \, || \, f_{\text{dino}}^j]; \Theta_{\text{edge}}\right)
\right),
\label{eq:pdgr_weight}
\end{equation}
where $[\cdot || \cdot]$ denotes channel-wise concatenation, $\mathcal{F}_{\text{MLP}}$ is a lightweight MLP with learnable parameters $\Theta_{\text{edge}}$, and $\zeta(\cdot)$ is the Softplus activation.

Let $H^{(0)} \in \mathbb{R}^{N \times C_{str}}$ denote the node-wise representation reshaped from $F_{str}$. Based on the constructed graph and the learned edge weights, we perform graph rectification as follows:
\begin{equation}
H_i^{(l+1)}=
\varphi\left(
\sum_{j\in\mathcal{N}(i)\cup\{i\}}
\frac{w_{ij}}{\sqrt{d_i d_j}}
\, H_j^{(l)} W^{(l)}
\right),
\label{eq:pdgr_gcn}
\end{equation}
where $H_i^{(l)}$ is the feature vector of node $i$ at layer $l$, $W^{(l)}$ is the learnable projection matrix, $d_i=\sum_j w_{ij}$ is the node degree, and $\mathcal{N}(i)$ denotes the neighbor set of node $i$. We use ReLU in the hidden layer and Layer Normalization in the output layer. After rectification, we obtain the rectified structure-dominated subspace $\hat{F}_{str}$.

Finally, $\hat{F}_{str}$ is recomposed with the preserved semantic subspace $F_{sem}$ according to the original channel indices to form the refined CLIP feature $F_{ref}$, which is then matched with the CLIP text embeddings to produce the refined correlation map for subsequent fusion.

\subsection{Uncertainty-Guided Adaptive Fusion}

After subspace decoupling and graph rectification, DR-Seg produces two branches with different roles: the original CLIP branch, which preserves robust language-aligned semantics, and the refined branch, which enhances local structure and boundary quality. These two branches should not be fused uniformly, because the relative importance of semantic reliability and structural correction varies across spatial regions. We therefore introduce an \textbf{Uncertainty-Guided Adaptive Fusion} module that adaptively balances the two branches based on pixel-wise prediction uncertainty.

Following existing methods~\cite{RSKT-seg, OVSNet}, we retain the multi-rotation CLIP branch to improve orientation robustness in RS scenes. In contrast, DINO-guided structural rectification is applied only to the canonical $0^\circ$ branch, since its role is to enhance visual structure rather than perform repeated orientation modeling, which also avoids the extra cost of constructing multiple DINO-guided graphs.

Specifically, for the rotated CLIP branch, the correlation map of the $i$-th rotation is computed as:
\begin{equation}
\mathbf{C}_{\text{clip}}^{i} = \frac{F_{\text{clip}}^{i} \cdot T_{\text{clip}}}{\|F_{\text{clip}}^{i}\| \, \|T_{\text{clip}}\|},
\label{eq:ugaf_clip_corr}
\end{equation}
and these aligned predictions are aggregated to yield the original semantic prediction branch $\mathbf{C}_{ori}$. For the refined branch, the correlation map is computed in the same normalized form:
\begin{equation}
\mathbf{C}_{\text{ref}} = \frac{F_{\text{ref}} \cdot T_{\text{clip}}}{\|F_{\text{ref}}\| \, \|T_{\text{clip}}\|}.
\label{eq:ugaf_ref_corr}
\end{equation}

To estimate where stronger structural correction is needed, we derive a pixel-wise uncertainty map from $\mathbf{C}_{ori}$. Let $p_k$ denote the softmax probability of class $k$ derived from $\mathbf{C}_{ori}$, and let $ U = -\sum_k p_k \log(p_k+\epsilon) $ be the corresponding entropy. We then define the normalized uncertainty gate as:
\begin{equation}
M_{unc}=\left(\frac{U}{\log(N_c)}\right)^2,
\label{eq:ugaf_uncertainty}
\end{equation}
where larger values indicate more ambiguous regions. Intuitively, semantically confident regions should remain dominated by the original CLIP branch, while uncertain regions should receive stronger structural correction. We therefore use $M_{unc}$ to modulate the refined branch:

\begin{equation}
\tilde{\mathbf{C}}_{ref} = \sigma(\psi(M_{unc})) \odot \mathbf{C}_{ref},
\label{eq:ugaf_modulated}
\end{equation}
where $\psi(\cdot)$ is a lightweight projection layer, $\sigma(\cdot)$ is the sigmoid activation, and $\odot$ denotes the Hadamard product.

\begin{table*}[t]
\centering
\caption{Comparison with state-of-the-art methods on the training dataset of DLRSD and iSAID. We report mIoU(\%) and mACC (\%) for every method. The best-performing results are presented in bold, while the second-best results are underlined.}
\label{tab:comparison}
\footnotesize
\setlength{\tabcolsep}{2.2pt}
\renewcommand{\arraystretch}{1.05}
\begin{tabular}{l l *{18}{c}}
\toprule
\multicolumn{20}{c}{\textbf{DLRSD as Training Dataset}} \\
\midrule
\multirow{2}{*}{\textbf{Method}} & \multirow{2}{*}{\textbf{Backbone}} & \multicolumn{2}{c}{\textbf{Potsdam}} & \multicolumn{2}{c}{\textbf{Vaihingen}} & \multicolumn{2}{c}{\textbf{UAVid}} & \multicolumn{2}{c}{\textbf{DLRSD}} & \multicolumn{2}{c}{\textbf{iSAID}} & \multicolumn{2}{c}{\textbf{LoveDA}} & \multicolumn{2}{c}{\textbf{UDD5}} & \multicolumn{2}{c}{\textbf{VDD}} & \multirow{2}{*}{\textbf{m-mIoU}} & \multirow{2}{*}{\textbf{m-mACC}} \\
\cmidrule(lr){3-4} \cmidrule(lr){5-6} \cmidrule(lr){7-8} \cmidrule(lr){9-10} \cmidrule(lr){11-12} \cmidrule(lr){13-14} \cmidrule(lr){15-16} \cmidrule(lr){17-18}
& & mIoU & mACC & mIoU & mACC & mIoU & mACC & mIoU & mACC & mIoU & mACC & mIoU & mACC & mIoU & mACC & mIoU & mACC & & \\
\midrule
SCAN\cite{scan} & ViT-B & 20.22 & 34.70 & 5.38 & 22.54 & 18.56 & 30.31 & 48.52 & 68.68 & 34.18 & 49.05 & 18.23 & 40.12 & 32.12 & 40.45 & 26.25 & 43.67 & 25.43 & 41.19 \\
SAN\cite{san} & ViT-B & 30.30 & 44.98 & 31.92 & 45.36 & 22.34 & 37.56 & 85.73 & 91.03 & 30.63 & 44.03 & 23.15 & 48.26 & 36.87 & 47.21 & 34.76 & 52.42 & 36.96 & 51.36 \\
SED\cite{sed} & ConvNeXt-B & 19.47 & 33.40 & 29.40 & 49.38 & 20.12 & 33.34 & 85.13 & 91.36 & 21.54 & 36.28 & 21.32 & 45.17 & 34.65 & 44.10 & 31.43 & 50.25 & 32.88 & 47.91 \\
Cat-Seg\cite{catseg} & ViT-B & 26.79 & 44.72 & 32.32 & 49.65 & 24.56 & 39.20 & 85.84 & 91.44 & 23.56 & 38.48 & 25.45 & 50.32 & 38.23 & 49.56 & 36.18 & 54.30 & 36.62 & 52.21 \\
OVRS\cite{OVSNet} & ViT-B & 27.47 & 42.07 & 33.71 & 49.01 & 25.23 & 40.18 & 85.98 & 91.52 & 39.09 & 54.43 & 28.67 & 52.10 & 39.10 & 50.65 & 37.34 & 55.15 & 39.57 & 54.51 \\
GSNet\cite{GSNet} & ViT-B & 26.46 & 43.20 & 35.15 & 52.62 & 25.42 & 40.70 & 84.12 & 90.53 & \underline{42.00} & \underline{59.19} & 29.32 & 53.02 & 40.05 & 51.72 & 38.10 & \underline{56.01} & 40.08 & 55.87 \\
RSKT-Seg\cite{RSKT-seg} & ViT-B & \underline{34.53} & \underline{50.71} & \underline{37.16} & \underline{54.81} & \underline{28.14} & \underline{42.86} & \underline{90.60} & \underline{94.89} & \textbf{44.04} & \textbf{61.23} & \textbf{32.49} & \underline{55.67} & \underline{42.99} & \underline{57.81} & \textbf{41.22} & \textbf{58.09} & \underline{43.90} & \underline{59.51} \\
DR-Seg (Ours) & ViT-B & \textbf{41.55} & \textbf{57.69} & \textbf{41.39} & \textbf{57.80} & \textbf{28.45} & \textbf{43.23} & \textbf{90.91} & \textbf{95.27} & 39.65 & 56.18 & \underline{32.38} & \textbf{56.33} & \textbf{46.36} & \textbf{62.14} & \underline{39.35} & 54.79 & \textbf{45.00} & \textbf{60.42} \\ 
\midrule
SCAN\cite{scan} & ViT-L & 27.45 & 39.22 & 15.23 & 29.45 & 20.28 & 34.43 & 52.42 & 72.43 & 44.28 & 67.25 & 23.17 & 35.36 & 34.14 & 43.25 & 29.24 & 45.57 & 30.78 & 45.87 \\
SAN\cite{san} & ViT-L & 37.25 & 46.28 & 39.22 & 48.33 & 23.53 & 38.14 & 86.45 & 91.25 & 49.56 & 67.25 & 25.33 & 37.54 & 37.23 & 48.45 & 35.83 & 53.25 & 41.80 & 53.81 \\
SED\cite{sed} & ConvNeXt-L & 29.35 & 37.95 & 39.02 & 58.62 & 21.33 & 35.64 & 87.68 & 91.24 & 51.23 & 68.24 & 24.55 & 36.83 & 35.73 & 45.15 & 32.53 & 51.34 & 40.18 & 53.13 \\
Cat-Seg\cite{catseg} & ViT-L & 35.75 & 49.03 & 42.30 & 60.65 & \underline{25.73} & \underline{40.54} & 88.68 & 93.34 & 53.34 & 70.86 & 28.64 & 38.73 & 40.24 & 51.65 & 39.14 & 55.85 & 44.23 & 57.58 \\
OVRS\cite{OVSNet} & ViT-L & 36.44 & 50.17 & 43.50 & \underline{63.31} & 24.13 & 34.83 & 88.85 & 93.64 & 52.65 & 69.59 & 31.53 & \underline{59.82} & 40.82 & 54.24 & 37.23 & 56.34 & 44.39 & 60.24 \\
GSNet\cite{GSNet} & ViT-L & \underline{37.85} & 52.35 & \underline{44.13} & 62.38 & 24.22 & 35.03 & 86.02 & 91.48 & 53.73 & 71.57 & \underline{32.52} & \textbf{60.23} & 40.92 & 57.04 & 37.34 & 57.04 & 44.56 & 60.89 \\
RSKT-Seg\cite{RSKT-seg} & ViT-L & 37.14 & \underline{53.75} & 42.29 & 63.03 & 23.88 & 38.75 & \textbf{91.58} & \textbf{95.41} & \underline{56.88} & \textbf{74.48} & 31.43 & 55.23 & \underline{45.73} & \underline{59.40} & \underline{40.55} & \underline{58.75} & \underline{46.18} & \underline{62.35} \\
DR-Seg (Ours) & ViT-L & \textbf{45.91} & \textbf{60.14} & \textbf{47.99} & \textbf{68.06} & \textbf{27.78} & \textbf{42.53} & \underline{91.13} & \underline{95.15} & \textbf{57.56} & \underline{73.92} & \textbf{34.10} & 56.59 & \textbf{46.63} & \textbf{62.61} & \textbf{41.00} & \textbf{60.16} & \textbf{49.01} & \textbf{64.90} \\
\midrule
\multicolumn{20}{c}{\textbf{iSAID as Training Dataset}} \\
\midrule
SCAN\cite{scan} & ViT-B & 18.25 & 33.17 & 8.72 & 27.20 & 9.87 & 15.30 & 16.09 & 38.25 & 62.34 & 76.48 & 12.56 & 32.10 & 8.45 & 28.60 & 15.32 & 35.10 & 18.95 & 35.78 \\
SAN\cite{san} & ViT-B & 14.82 & 34.84 & 16.23 & 34.38 & 12.32 & 18.32 & 18.82 & 42.36 & 85.43 & 90.36 & 18.45 & 40.25 & 12.11 & 32.45 & 22.10 & 40.25 & 25.04 & 41.65 \\
SED\cite{sed} & ConvNeXt-B & 5.78 & 17.52 & 9.36 & 21.62 & 14.80 & 18.90 & 21.48 & 45.15 & 93.31 & 96.66 & 20.10 & 42.30 & 10.23 & 31.50 & 17.32 & 38.12 & 24.05 & 38.97 \\
Cat-Seg\cite{catseg} & ViT-B & 15.23 & 37.17 & 14.03 & 38.61 & 15.47 & 23.57 & 20.41 & 44.08 & \underline{94.16} & \underline{96.72} & 23.50 & 41.55 & 12.10 & 38.55 & 19.62 & 51.38 & 26.82 & 46.45 \\
OVRS\cite{OVSNet} & ViT-B & 15.57 & 38.94 & 14.66 & 38.68 & 16.22 & 24.05 & 21.06 & 45.48 & \textbf{94.60} & \textbf{96.87} & 25.30 & 42.32 & 11.90 & 38.80 & 21.42 & 51.25 & 27.59 & 47.05 \\
GSNet\cite{GSNet} & ViT-B & 15.12 & 36.16 & 14.25 & 41.15 & 15.93 & 24.33 & \underline{26.20} & \underline{57.07} & 90.00 & 93.60 & 26.80 & 42.84 & 12.44 & 39.62 & 22.22 & 42.33 & 27.87 & 47.14 \\
RSKT-Seg\cite{RSKT-seg} & ViT-B & \textbf{20.28} & \textbf{46.71} & \underline{17.47} & \textbf{50.84} & \underline{17.15} & \underline{34.78} & 24.80 & 55.94 & 93.16 & 96.37 & \underline{28.07} & \underline{46.58} & \underline{13.94} & \underline{43.47} & \underline{25.34} & \underline{47.18} & \underline{30.03} & \underline{52.73} \\
DR-Seg (Ours) & ViT-B & \underline{19.40} & \underline{41.24} & \textbf{19.66} & \underline{45.34} & \textbf{18.73} & \textbf{36.46} & \textbf{28.84} & \textbf{59.22} & 92.17 & 95.56 & \textbf{28.97} & \textbf{52.74} & \textbf{31.12} & \textbf{58.44} & \textbf{32.20} & \textbf{53.41} & \textbf{33.89} & \textbf{55.30} \\
\midrule
SCAN\cite{scan} & ViT-L & 28.32 & 52.47 & 14.23 & 34.25 & 14.32 & 26.00 & 21.44 & 53.26 & 64.28 & 85.46 & 18.50 & 38.20 & 16.22 & 32.51 & 20.18 & 36.70 & 24.69 & 44.86 \\
SAN\cite{san} & ViT-L & 24.72 & 56.54 & 22.49 & 50.78 & 15.01 & 26.50 & 20.54 & 49.32 & 87.22 & 92.54 & 22.33 & 44.22 & 21.35 & 37.66 & 26.42 & 42.30 & 30.01 & 49.98 \\
SED\cite{sed} & ConvNeXt-L & 11.85 & 23.87 & 12.61 & 25.73 & 15.21 & 26.80 & 23.80 & 50.36 & 94.32 & 96.84 & 23.24 & 45.13 & 22.42 & 39.15 & 28.50 & 44.20 & 28.99 & 44.01 \\
Cat-Seg\cite{catseg} & ViT-L & 23.90 & 49.49 & 21.74 & 51.25 & 16.10 & 26.90 & 28.80 & 59.56 & \underline{94.77} & 96.96 & 25.11 & 48.25 & 24.32 & 42.77 & 30.16 & 47.20 & 33.11 & 52.80 \\
OVRS\cite{OVSNet} & ViT-L & 26.39 & 50.15 & \underline{28.80} & \underline{54.20} & 16.24 & 27.00 & \textbf{32.25} & 60.35 & \textbf{94.86} & \textbf{97.06} & 27.98 & 49.01 & 31.88 & 55.32 & 31.01 & 55.30 & 36.18 & 56.05 \\
GSNet\cite{GSNet} & ViT-L & 28.50 & \underline{52.00} & 25.10 & 52.80 & 15.98 & 27.10 & 31.50 & \underline{63.05} & 93.11 & 95.98 & 27.21 & 49.53 & 32.24 & 56.20 & 32.07 & 55.21 & 35.71 & 56.48 \\
RSKT-Seg\cite{RSKT-seg} & ViT-L & \underline{28.57} & 51.69 & 25.55 & 52.66 & \underline{17.36} & \underline{35.75} & 31.57 & 61.66 & 93.96 & 96.63 & \underline{29.62} & \underline{51.01} & \textbf{34.01} & \underline{57.38} & \underline{33.40} & \underline{56.50} & \underline{36.76} & \underline{57.91} \\
DR-Seg (Ours) & ViT-L & \textbf{29.67} & \textbf{52.90} & \textbf{28.89} & \textbf{56.55} & \textbf{18.90} & \textbf{36.98} & \underline{31.98} & \textbf{64.06} & 94.61 & \underline{97.02} & \textbf{30.52} & \textbf{53.56} & \underline{33.42} & \textbf{59.82} & \textbf{34.75} & \textbf{60.23} & \textbf{37.84} & \textbf{60.14} \\
\bottomrule
\end{tabular}
\Description{Comprehensive comparison of open-vocabulary segmentation methods on multiple remote sensing datasets. Results are shown for both DLRSD and iSAID as training datasets, with ViT-B and ViT-L backbones. Evaluation metrics include mean Intersection over Union (mIoU) and mean Accuracy (mACC) across eight test datasets: Potsdam, Vaihingen, UAVid, DLRSD, iSAID, LoveDA, UDD5, and VDD. Best results are highlighted in bold, second-best underlined.}
\end{table*}

After obtaining the uncertainty-modulated refined correlation map $\tilde{\mathbf{C}}_{ref}$, we project both $\tilde{\mathbf{C}}_{ref}$ and the original correlation map $\mathbf{C}_{ori}$ into a latent embedding space. The original branch mainly provides reliable semantic embeddings, while the refined branch is further processed by a lightweight multi-scale convolutional block to capture both local structural corrections and broader contextual cues. The two branches are then fused by a lightweight fusion head, and the fused representation is fed into the subsequent aggregation layers and decoder for final segmentation prediction.

\section{Experiments}
\subsection{Experimental Setup}

\textbf{Dataset.} Following RSKT-Seg~\cite{RSKT-seg}, we evaluate on eight remote sensing segmentation benchmarks: DLRSD~\cite{dlrsd}, iSAID~\cite{isaid}, Potsdam, Vaihingen, LoveDA~\cite{loveda}, UAVid~\cite{uavid}, UDD5~\cite{udd5}, and VDD~\cite{vdd}. These datasets cover satellite, airborne, and UAV imagery with diverse spatial resolutions, scene layouts, and semantic granularity. We train on DLRSD (17 categories) and iSAID (15 categories) based on the standard protocol, and evaluate on all eight datasets to assess both in-domain and cross-dataset generalization under variations in object scale, imaging viewpoint, scene complexity, and class composition.

\noindent \textbf{Implementation Details.}
Our method is implemented in PyTorch based on Detectron2~\cite{detectron2}. We use CLIP ViT-B/16 and ViT-L/14 as the backbone, and adopt an RS-pretrained DINO encoder as the structural prior. Unless otherwise specified, RS-DINOv3~\cite{dinov3} ViT-L/16 is used in the default setting. For all categories, we use the text prompt template ``A photo of a [CLASS] in the scene''. The input resolution is set to $384\times384$ for ViT-B/16 and $336\times336$ for ViT-L/14. Following the benchmark protocol of RSKT-Seg~\cite{RSKT-seg}, we train the model for 30K iterations using AdamW with a batch size of 8, weight decay of $1\times10^{-4}$, and an initial learning rate of $2\times10^{-4}$. Unless otherwise stated, the key hyperparameters are $\lambda=0.3$, $\rho=0.5$, $\sigma_f=2.0$, $\sigma_s=0.05$, and $k=75$. All experiments are conducted on two NVIDIA RTX 5090 GPUs with 32GB of memory each. We report mIoU, mACC, and fwIoU, where mIoU serves as the primary metric for class-balanced evaluation, mACC reflects the average per-class accuracy, and fwIoU emphasizes overall performance by accounting for class frequency.

\subsection{Comparison Results}

Table~\ref{tab:comparison} compares DR-Seg with recent open-vocabulary segmentation methods under the standard benchmark protocol. To disentangle the effect of the structural prior from that of the proposed design, we further compare different methods under matched priors in Table~\ref{tab:prior}. Overall, DR-Seg achieves the best average performance in nearly all settings, demonstrating strong open-vocabulary segmentation capability and robust generalization across diverse remote sensing domains. This advantage is consistent across both training protocols and backbone scales. When trained on DLRSD, the gain is particularly pronounced: with ViT-L, DR-Seg reaches \textbf{49.01\%} m-mIoU and \textbf{64.90\%} m-mACC, surpassing RSKT-Seg by \textbf{+2.83} and \textbf{+2.55} points, respectively. When trained on iSAID, DR-Seg again achieves the best average performance, reaching \textbf{33.89\%/55.30\%} m-mIoU/m-mACC with ViT-B and \textbf{37.84\%/60.14\%} with ViT-L. The gains are especially evident on cross-dataset benchmarks such as UDD5 and VDD, suggesting that selective structural enhancement improves spatial precision while preserving CLIP's semantic generalization under domain shift.

\begin{figure}[!t]
\centering
\includegraphics[width=\linewidth]{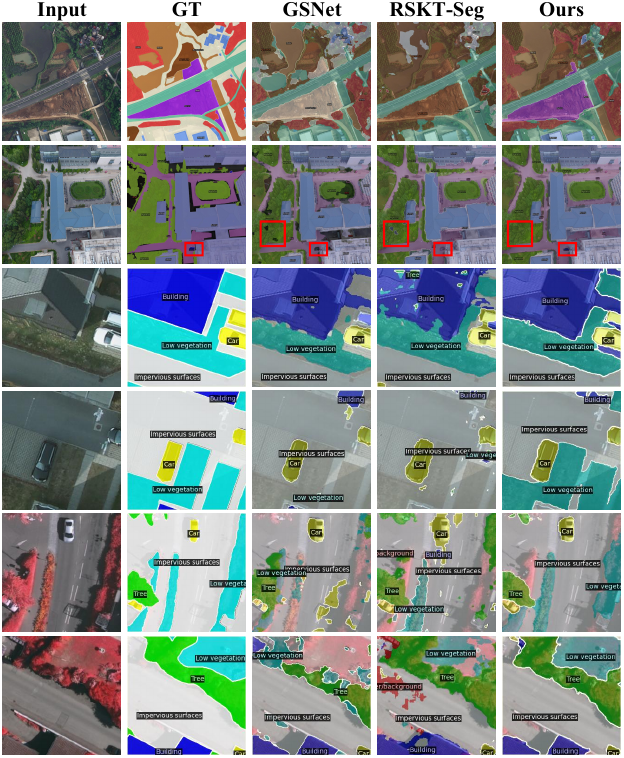}
\caption{Qualitative comparison with GSNet and RSKT-Seg. DR-Seg yields more accurate boundaries, cleaner category assignments, and better recovery of small or thin structures. The ViT-L model trained on DLRSD produces the results.}
\label{fig:visual}
\Description{Qualitative comparison of segmentation results across five methods. The figure contains five rows showing: (1) input remote sensing images, (2) ground truth segmentation masks, (3) GSNet predictions, (4) RSKT-Seg predictions, and (5) our DR-Seg predictions. Our method produces segmentation masks that more closely match the ground truth, with fewer misclassifications and better boundary delineation compared to GSNet and RSKT-Seg. The model uses the ViT-L backbone trained on the DLRSD dataset.}
\end{figure}

Figure~\ref{fig:visual} presents qualitative comparisons. Compared with GSNet and RSKT-Seg, DR-Seg yields cleaner boundaries, fewer category confusions, and more complete predictions for small or thin structures. These improvements are particularly visible along building edges and on small cars, where previous methods tend to produce discontinuous or coarsely delineated predictions. Such visual results explicitly validate the design motivation of our method.

\subsection{Ablation Study}
We conduct a series of ablation studies to examine the effectiveness of the proposed components and the impact of key design choices. Unless otherwise specified, all ablation experiments are trained on DLRSD, use CLIP ViT-L as the backbone, and are evaluated by averaging the results on four datasets: Potsdam, VDD, UAVid, and UDD5.

\noindent \textbf{Component Analysis.}
Table~\ref{tab:ablation_components} evaluates the contribution of PDGR, SPSD, and UGAF, where all variants are built upon the same baseline. Adding PDGR alone improves m-mIoU from 34.15\% to 36.76\%, indicating that DINO-guided graph rectification effectively introduces valuable structural priors. Building on this, when SPSD is incorporated to decouple the CLIP features before the rectification process, the performance further increases to 37.97\%. This demonstrates a strong synergy between the two modules: SPSD prepares the feature space by isolating the structure-dominated subspace, which enables PDGR to perform highly targeted structural enhancement while safely preserving language-aligned semantics. Alternatively, applying UGAF alongside PDGR yields 38.38\%, confirming the benefit of adaptive fusion in balancing the original and refined branches. Finally, integrating all three components achieves the best performance of \textbf{40.33\%} m-mIoU, \textbf{52.09\%} m-fwIoU, and \textbf{56.36\%} m-mACC, improving over the baseline by \textbf{+6.18}, \textbf{+7.53}, and \textbf{+5.88} points, respectively. These results show that SPSD, PDGR, and UGAF are highly complementary and jointly contribute to the final performance.

\begin{table}[!t]
\centering
\caption{Effectiveness of the proposed components. All variants are built upon the same baseline. The best-performing results are presented in bold.}
\label{tab:ablation_components}
\setlength{\tabcolsep}{5pt}
\begin{tabular}{ccc|ccc}
\toprule
PDGR & SPSD & UGAF & m-mIoU & m-fwIoU & m-mACC \\
\midrule
 &  &  & 34.15 & 44.56 & 50.48 \\
\checkmark &  &  & 36.76 & 47.88 & 51.78 \\
\checkmark & \checkmark &  & 37.97 & 48.80 & 53.65 \\
\checkmark &  & \checkmark & 38.38 & 51.86 & 55.25 \\
\checkmark & \checkmark & \checkmark & \textbf{40.33} & \textbf{52.09} & \textbf{56.36} \\
\bottomrule
\end{tabular}
\Description{Ablation of the proposed modules PDGR, SPSD, and UGAF. All variants are built upon the same baseline. The full model achieves the best performance with 40.33\% m-mIoU, 52.09\% m-fwIoU, and 56.36\% m-mACC.}
\end{table}

\begin{figure}[!t]
    \centering
    \includegraphics[width=\linewidth]{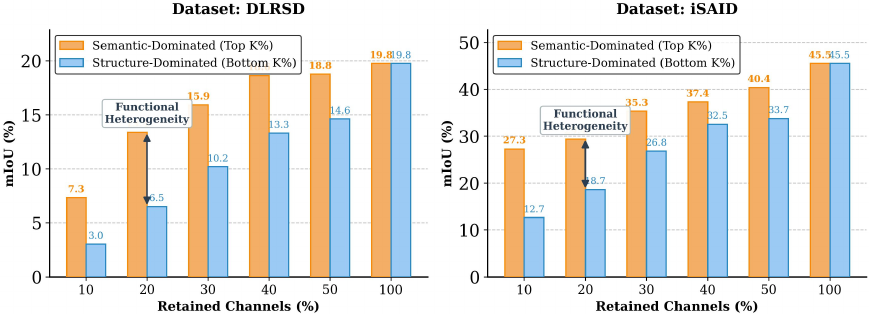}
    \caption{Zero-shot segmentation performance of the top-ranked and bottom-ranked CLIP channel groups under varying retention ratios. Channels are strictly sorted by our proposed semantic selectivity criterion and evaluated independently using a frozen CLIP encoder. This significant gap reveals distinct functional heterogeneity, validating our semantics-preserving decoupling strategy.}
    \label{fig:decouple}
\end{figure}

\noindent \textbf{Verification of SPSD.}
To further validate the proposed semantic-preserving subspace decoupling, Fig.~\ref{fig:decouple} compares the \textit{zero-shot segmentation} performance obtained by retaining different proportions of the top-ranked and bottom-ranked channels from the frozen CLIP encoder. A clear and consistent performance gap is observed across different retention ratios, revealing distinct functional heterogeneity of CLIP channels. The top-ranked channels exhibit substantially stronger language-aligned segmentation capability, whereas the bottom-ranked channels contribute significantly less to semantic prediction. This confirms that the proposed ranking effectively isolates semantically critical channels. Furthermore, to demonstrate that the bottom-ranked channels capture dense, low-level structural responses rather than mere noise, additional analyses of their spatial activation sparsity are provided in the supplementary material.

\begin{figure}[!t]
    \centering
    \includegraphics[width=\linewidth]{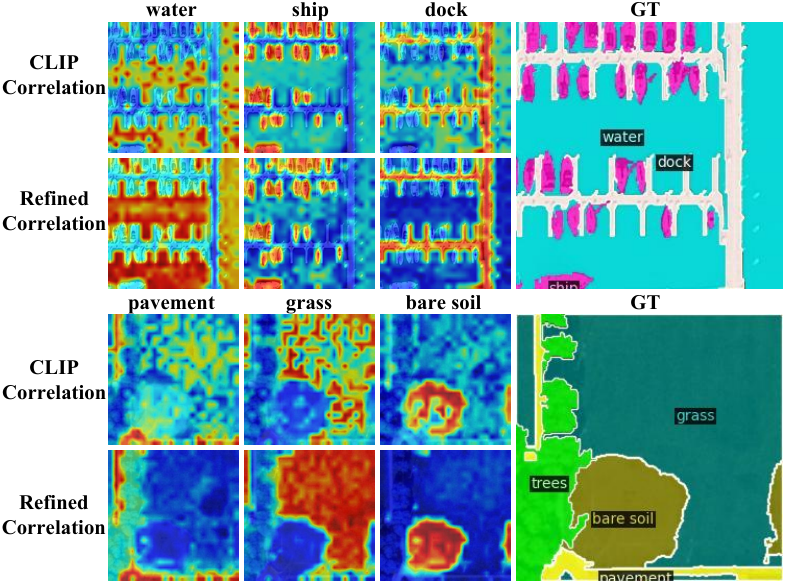}
    \caption{Visualization of class-wise correlation maps from the original CLIP branch and the refined branch in DR-Seg.}
    \label{fig:refined}
\end{figure}

\begin{table}[!t]
\centering
\caption{Impact of structural priors.}
\label{tab:prior}
\setlength{\tabcolsep}{4pt}
\begin{tabular}{l|c|ccc}
\toprule
Method & Prior & m-mIoU & m-fwIoU & m-mACC \\
\midrule
\multicolumn{5}{c}{\textit{Matched prior comparison}} \\
\midrule
GSNet & RS-DINOv1 & 35.08 & -- & 50.37 \\
RSKT-Seg & RS-DINOv1 & 36.83 & 48.52 & 52.66 \\
\textbf{DR-Seg} & \textbf{RS-DINOv1} & \textbf{38.37} & \textbf{50.94} & \textbf{55.56} \\
\midrule
\multicolumn{5}{c}{\textit{Scaling to stronger prior}} \\
\midrule
GSNet & RS-DINOv3 & 38.30 & 48.69 & 51.82 \\
RSKT-Seg & RS-DINOv3 & 36.39 & 47.67 & 51.02 \\
\textbf{DR-Seg} & \textbf{RS-DINOv3} & \textbf{40.33} & \textbf{52.09} & \textbf{56.36} \\
\bottomrule
\end{tabular}
\Description{Effect of structural priors. Under the same RS-DINOv1 prior, DR-Seg outperforms previous methods. With the stronger RS-DINOv3 prior, DR-Seg achieves the best overall performance.}
\end{table}

\noindent \textbf{Verification of PDGR.}
Fig.~\ref{fig:refined} further visualizes the effect of the proposed structural rectification by comparing the class-wise correlation maps from the original CLIP branch and the refined branch. Compared with the original CLIP correlation, the refined branch exhibits more spatially concentrated responses and clearer structural contours, especially around object boundaries and small regions. This observation provides direct evidence that PDGR improves structural localization in a targeted manner and complements the preserved semantic prediction from the original CLIP branch. More visual examples are provided in the supplementary material.

\noindent \textbf{Impact of Structural Priors.}
Table~\ref{tab:prior} examines the role of structural priors in our framework. Under the same RS-DINOv1 prior, DR-Seg improves the m-mIoU from 36.83\% (RSKT-Seg) to 38.37\%, along with consistent gains in m-fwIoU and m-mACC. This demonstrates that our performance improvements do not stem merely from using a stronger encoder. Furthermore, when adopting the more advanced RS-DINOv3 prior, DR-Seg achieves further gains, reaching peak scores of \textbf{40.33\%} in m-mIoU and \textbf{56.36\%} in m-mACC. These results confirm that the proposed decouple-and-rectify design is not only highly effective under matched priors but also scales well with stronger structural guidance.

\noindent \textbf{Effect of Fusion Strategy.}
Table~\ref{tab:fusion} compares different strategies for combining the original CLIP branch and the rectified branch. Simple mean fusion already performs reasonably well, suggesting that the rectified branch provides complementary cues. However, naive concatenation and the separate strategy are less effective. Our uncertainty-guided adaptive fusion achieves the best results across all metrics, confirming the importance of spatially adaptive fusion for balancing semantic reliability and structural correction. This also shows that the two branches should be fused selectively rather than treated as uniformly reliable across all spatial regions.

\begin{table}[!t]
\centering
\caption{Comparison of fusion strategies.}
\label{tab:fusion}
\setlength{\tabcolsep}{6pt}
\begin{tabular}{lccc}
\toprule
Fusion & m-mIoU & m-fwIoU & m-mACC \\
\midrule
mean & 39.62 & 51.66 & 54.96 \\
concat & 37.21 & 49.79 & 53.02 \\
separate & 36.55 & 47.29 & 51.34 \\
\textbf{UGAF (ours)} & \textbf{40.33} & \textbf{52.09} & \textbf{56.36} \\
\bottomrule
\end{tabular}
\Description{Comparison of fusion strategies for combining the original CLIP branch and the rectified branch. UGAF achieves the best performance across all metrics.}
\end{table}
\begin{table}[!t]
\centering
\caption{Effect of semantic ratio $\rho$.}
\label{tab:rho}
\setlength{\tabcolsep}{8pt}
\begin{tabular}{cccc}
\toprule
$\rho$ & m-mIoU & m-fwIoU & m-mACC \\
\midrule
0.3 & 37.71 & 51.34 & 54.62 \\
0.4 & 38.80 & 51.13 & 54.79 \\
\textbf{0.5} & \textbf{40.33} & \textbf{52.09} & \textbf{56.36} \\
0.6 & 39.25 & 52.00 & 54.52 \\
0.7 & 37.75 & 49.37 & 52.68 \\
\bottomrule
\end{tabular}
\Description{Effect of semantic ratio $\rho$. The best performance is achieved at $\rho=0.5$.}
\end{table}
\begin{table}[!t]
\centering
\caption{Effect of graph sparsity parameter $k$.}
\label{tab:topk}
\setlength{\tabcolsep}{8pt}
\begin{tabular}{cccc}
\toprule
Top-$k$ & m-mIoU & m-fwIoU & m-mACC \\
\midrule
20  & 37.57 & 50.62 & 55.55 \\
40  & 38.51 & 50.27 & 54.86 \\
50  & 38.95 & 51.74 & 55.60 \\
\textbf{75 } & \textbf{40.33} & \textbf{52.09} & \textbf{56.36} \\
100 & 38.05 & 49.26 & 53.00 \\
\bottomrule
\end{tabular}
\Description{Effect of the graph sparsity parameter $k$ in PDGR. The best performance is achieved at $k=75$.}
\end{table}

\noindent \textbf{Impact of Semantic Ratio $\rho$.} 
Table~\ref{tab:rho} studies the semantic ratio $\rho$ in SPSD. The best performance is achieved at $\rho=0.5$, indicating that a balanced partition between semantic preservation and structural rectification is most effective. This suggests that allocating too many channels to either subspace weakens the intended balance between semantic fidelity and structural enhancement.

\noindent \textbf{Impact of Top-$k$ Selection.}
Table~\ref{tab:topk} studies the graph sparsity parameter $k$ used in PDGR. Performance first improves as $k$ increases and peaks at $k=75$, indicating that a moderately sparse graph best balances structural completeness and noise suppression. When $k$ is too small, important structural relations may be missed, whereas overly dense graphs introduce irrelevant propagation across regions. More hyperparameter analyses are provided in the supplementary material.

\section{Conclusion}

In this work, we introduce DR-Seg, a decouple-and-rectify framework for open-vocabulary remote sensing segmentation. Motivated by our analysis that CLIP features exhibit functional heterogeneity across channels, DR-Seg performs semantics-preserving structural enhancement. Specifically, it decouples CLIP representations into semantics-dominated and structure-dominated subspaces, rectifies structural priors under DINO guidance, and adaptively fuses the original CLIP branch with the refined branch. Extensive experiments on eight remote sensing benchmarks demonstrate state-of-the-art performance and strong cross-dataset generalization, suggesting that semantics-preserving structural enhancement is a more effective alternative to holistic fusion for open-vocabulary remote sensing segmentation.

\bibliographystyle{ACM-Reference-Format}
\bibliography{my_paper/inference}

\end{document}